\documentclass{article}

\usepackage[preprint]{neurips_2019}

\usepackage[utf8]{inputenc} % allow utf-8 input
\usepackage[T1]{fontenc}    % use 8-bit T1 fonts
\usepackage{hyperref}       % hyperlinks
\usepackage{url}            % simple URL typesetting
\usepackage{booktabs}       % professional-quality tables
\usepackage{amsfonts}       % blackboard math symbols
\usepackage{nicefrac}       % compact symbols for 1/2, etc.
\usepackage{microtype}      % microtypography
\usepackage{amsmath}
\usepackage{graphicx}
\usepackage{subcaption}
\usepackage{wrapfig}

\newcommand\eglanz{Emily Glanz}
\newcommand\mcmahan{Brendan McMahan}
\newcommand\harda{Andrew Hard}
\newcommand\assistant{Google Assistant}
\newcommand\googleai{Google AI}

\title{Federated Evaluation of On-device Personalization}

\author{Kangkang Wang, Rajiv Mathews, Chlo{\'{e}} Kiddon, Hubert Eichner, Fran{\c{c}}oise Beaufays, Daniel Ramage\\
  Google, Inc.\\
  {\texttt \{kangkang, mathews, loeki, huberte, fsb, dramage\}@google.com} \\}

\author{%
  Kangkang Wang\\
  Google \\
  \texttt{kangkang@google.com} \\
  \And
  Rajiv Mathews \\
  Google \\
  \texttt{mathews@google.com} \\
  \AND
  Chlo{\'{e}} Kiddon \\
  Google \\
  \texttt{loeki@google.com} \\
  \And
  Hubert Eichner \\
  Google \\
  \texttt{huberte@google.com} \\
  \And
  Fran{\c{c}}oise Beaufays \\
  Google \\
  \texttt{fsb@google.com} \\
  \And
  Daniel Ramage \\
  Google \\
  \texttt{dramage@google.com} \\
}

\begin{document}

\maketitle

\begin{abstract}
  Federated learning is a distributed, on-device computation framework
  that enables training global models without exporting sensitive user
  data to servers. In this work, we describe methods to extend the
  federation framework to evaluate strategies for personalization of
  global models. We present tools to analyze the effects of
  personalization and evaluate conditions under which personalization
  yields desirable models. We report on our experiments personalizing
  a language model for a virtual keyboard for smartphones with a
  population of tens of millions of users. We show that a significant
  fraction of users benefit from personalization.
\end{abstract}

\section{Introduction}

As users increasingly shift to mobile devices as their primary
computing device~\citep{pewinternet}, we hypothesize that the
information on devices allows for personalizing global models to
better suit the needs of individual users. This can be achieved in a
privacy-preserving way by fine-tuning a global model using standard
optimization methods on data stored locally on a single device.  While
we expect personalization to be beneficial for most users, we need to
make sure it doesn't make things worse for some users, e.g. by
overfitting.

In this paper, we describe extensions to the Federated
Learning~\citep{bonawitz2019towards} (FL) framework for evaluating the
personalization of global models. We study this using an RNN language
model for the keyboard next-word prediction
task~\citep{hard2018federated}. We show that we can derive and impose
conditions under which a personalized model is deployed if and only if
it makes the user's experience better. We further show that it is
possible to personalize models that benefit a significant fraction of
users.

\section{Federated Personalization Evaluation}

Federated Learning is a distributed model training paradigm where data
never leaves users' devices. Only minimal and ephemeral updates to the
model are transmitted by the clients to the server where they are
aggregated into a single update to the global
model~\citep{fedlearn}. FL can be further combined with other
privacy-preserving techniques like secure multi-party
computation~\citep{Bonawitz2017} and differential
privacy~\citep{McMahan2017LearningDP, agarwal2018cpsgd,
  abadi2016deep}. \citet{hard2018federated} showed that FL can be used
to train an RNN language model that outperforms an identical model
trained using traditional server-side techniques, when evaluated on
the keyboard next-word prediction task.

\begin{wrapfigure}[23]{R}{0.5\textwidth}
  \centering
  \includegraphics[width=0.48\textwidth]{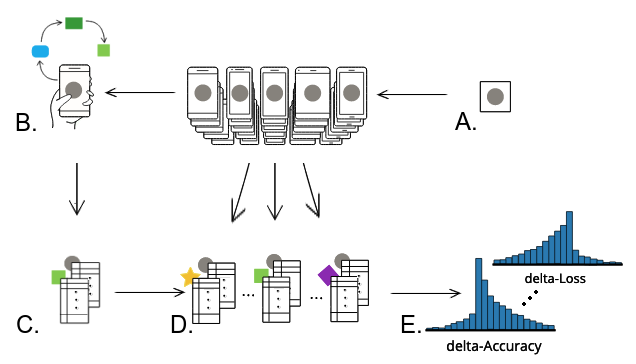}
  \caption{An illustration of \textit{Federated Personalization Evaluation}: (A)
    the global model (gray circle) is sent to client devices, (B) the device
    computes SGD updates on the train partition of the local data, resulting in
    a personalized model (the green square), (C) the device computes a report of
    metrics for the global and personalized model on the test partition of the
    local data, (D) pairs of metric reports are sent by various devices to the
    server, (E) the server computes histograms of various delta metrics.}
  \label{fig:personalization_eval_illustration}
\end{wrapfigure}

Such a global model is necessarily a consensus model, and it stands to
reason that population-wide accuracy can be further improved through
personalization on individual users' data. However, such on-device
refinements cannot be tested server-side because the training/eval
data is not collected centrally.  It is reasonable to expect that,
given the nature of neural network training, personalizing models
might make the experience of some users worse. We will show that we
can prevent such undesired effects by carefully calibrating the model
hyperparameters, and by building a gating mechanism that accepts or
rejects personalized models for use in inference.

In this paper, we introduce an extension to the FL framework for
evaluating personalization accuracy and for determining the training
and acceptance hyperparameters - \textit{Federated Personalization
  Evaluation} (FPE). As in the FL setting, mobile phones connect to a
server when idle, charging and on an unmetered network
\citep{bonawitz2019towards}. Selected devices are served a baseline
model along with instructions on how to train it using the device's
dataset in the form of a TensorFlow graph~\citep{tensorflow}. In FL,
the device would compute and send its model update to the server for
aggregation, but in FPE, the device instead does
five steps: it splits the local on-device dataset into a train and
test partition using practitioner-defined criteria; it computes
metrics of the baseline model on the test data; it fine-tunes the
model on the training set; it computes metrics of the personalized
model on the test set; and finally it computes and uploads the change
in metrics between the personalized and baseline variants.  The server
aggregates the metrics it receives from various clients to compute
histograms of various delta metrics.

Figure~\ref{fig:personalization_eval_illustration} illustrates this
process.  FPE allows us to evaluate the benefit of personalization and
identify good hyperparameters using the existing infrastructure for
federated learning, without any user visible impact. These conclusions
can then be used for \textit{live} inference using personalized
models, though live inference is beyond the scope of this paper.

\section{Method}

\subsection{Network Architecture}

The network architecture of the next-word prediction model is
described in ~\citet{hard2018federated}. We use a variant of the Long
Short-Term Memory (LSTM)~\citep{lstm} recurrent neural network called
the Coupled Input and Forget Gate (CIFG)~\citep{cifg}. The input
embedding and output projection matrices are tied to reduce the model
size~\citep{shareioemb,tyingioemb}. For a vocabulary of size $V$, a
one-hot encoding $v \in {\mathbb{R}}^{V}$ is mapped to a dense
embedding vector $d \in {\mathbb{R}}^{D}$ by $d = W v$ with an
embedding matrix $W \in {\mathbb{R}}^{D \times V}$. The output
projection of the CIFG, also in ${\mathbb{R}}^{D}$, is mapped to the
output vector ${W}^{\mathsf{T}} h \in {\mathbb{R}}^{V}$. A softmax
layer converts the raw logits into normalized
probabilities. Cross-entropy loss over the output and target labels is
used for training. We use a vocabulary of $V=\text{10,000}$ words,
including the special beginning-of-sentence, end-of-sentence, and
out-of-vocabulary tokens. The input embedding and CIFG output
projection dimension $D$ is set to 96. A single layer CIFG with 670
units is used. The network has 1.4 million parameters.

\subsection{Global Model Training}
\label{sec:global_model_training}

The next-word prediction model is trained using FL on a population of
users whose language is set to US English, as described in
~\citet{hard2018federated}. The \texttt{FederatedAveraging}
algorithm~\citep{fedlearn} is used to aggregate distributed client SGD
updates. Training progresses synchronously in ``rounds''. Every
client, indexed by $k$, participating in a given round, indexed by
$t$, computes the average gradient, ${g}_{k}$, on its local data
${n}_{k}$, with the current model ${w}_{t}$ using stochastic gradient
descent (SGD). For a client learning rate $\epsilon$, the local client
update, ${w}_{t+1}^{k}$, is given by ${w}_{t} - \epsilon {g}_{k}
\rightarrow {w}_{t+1}^{k}$. The server performs a weighted aggregation
of the client models to obtain a new global model, ${w}_{t+1}$:
$\sum_{k=1}^{K} \frac{{n}_{k}}{N} {w}_{t+1}^{k} \rightarrow
{w}_{t+1}$, where $N = \sum_{k} {n}_{k}$. The server update is
achieved via the Momentum optimizer, using Nesterov accelerated
gradient~\citep{nesterov, pmlr-v28-sutskever13}, a momentum
hyperparameter of 0.9, and a server learning rate of 1.0. Training
converges after 3000 training rounds, over the course of which 600
million sentences are processed by 1.5 million clients. Training
typically takes 4 to 5 days.

\subsection{Model Personalization Strategies}

A personalization strategy consists of the model graph, the initial
parameter values, and the training hyperparameters - client learning
rate, train batch size, and stopping criteria. Throughout our experiments,
the model graph and initial parameter values are set to be the federated
trained next word prediction model described in
Section~\ref{sec:global_model_training}. The effect of the personalization
learning on the model is evaluated via various training hyperparameters.

Given a personalization strategy, the personalized model can then be
trained from the initial global model using individual client’s
training cache data.  Data are cached on mobile devices on which the
language is set to US English.  During a training process, the client
data first gets split into train and test partitions (80\% and 20\%
based on the temporal order). Stochastic gradient descent
is used for model training with specified learning rate ($L$) and batch
size ($B$). The stopping criteria are based on number of tokens ($T$)
observed and number of epochs ($E$) trained. The training process stops
when one of the criteria is satisfied.

\section{Experiments}

The performance of the personalized model is evaluated using the
prediction accuracy metric, defined as the ratio of the number of
correct predictions to the total number of tokens.

%% \begin{wraptable}{L}{0.4\textwidth}
\begin{wraptable}{l}{0.5\textwidth}
  \centering
  \begin{tabular}{l c c}
    \toprule
        {\small \textit{Hyperparameters}} & {\small \textit{Accuracy delta}}  \\
        \midrule
        $B=5, L=0.01$  & 0.012 \\
        $B=5, L=0.1$   & 0.024 \\
        $B=5, L=1.0$   & -0.019 \\
        \midrule
        $B=10, L=0.01$ & 0.008 \\
        $B=10, L=0.1$  & 0.022 \\
        $B=10, L=1.0$  & 0.002 \\
        \midrule
        $B=20, L=0.01$ & 0.005 \\
        $B=20, L=0.1$  & 0.018 \\
        $B=20, L=1.0$  & 0.015 \\
        \bottomrule
  \end{tabular}
  \caption{The results from personalization eval experiments. Metrics are
    reported from over 500,000 clients. Mean of baseline accuracy is
    $0.166\pm0.001$.}
  \label{tab:table1}
\end{wraptable}

Experiments are conducted to study the influence of client train batch
size and learning rate on the personalization performance. The
stopping criteria are set to $T \geq 5000$ or $E \geq1$. We wish to
assess the benefits brought by personalization across users. However,
each user experiences a different baseline accuracy, depending on
their style, use of language register, etc. Therefore, it makes sense
to measure the difference between prediction accuracies before and
after personalization for each user, and observe the distribution of
these differences in addition to their average. The histograms report
only prediction-accuracy metrics and are computed over tens of
thousands of users. Metric reports including the baseline accuracies
and the personalized model accuracies from over 300,000 users' devices
are received. The delta metrics between the personalized model and the
baseline model are summarized in Table~\ref{tab:table1}.

\begin{figure*}
  \centering
  \begin{subfigure}{.32\linewidth}
    \includegraphics[width=\columnwidth]{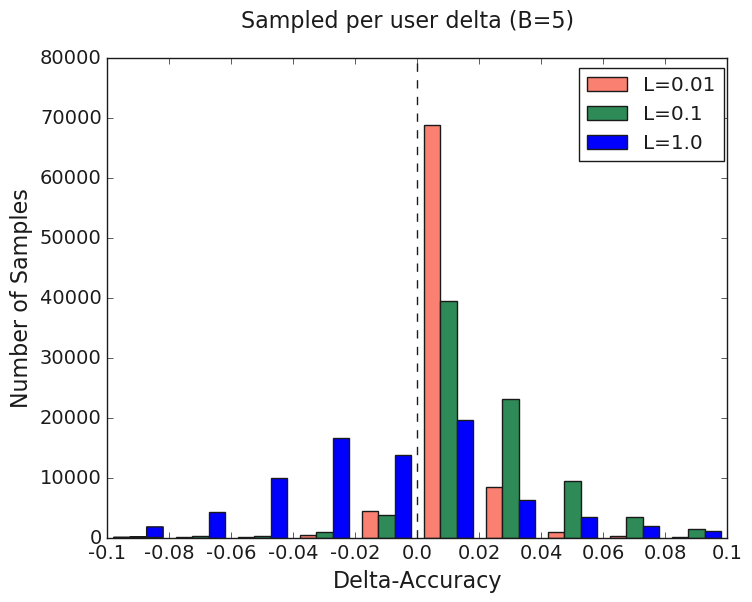}
    \caption{}
    \label{fig:1a}
  \end{subfigure}
  \begin{subfigure}{.32\linewidth}
    \includegraphics[width=\columnwidth]{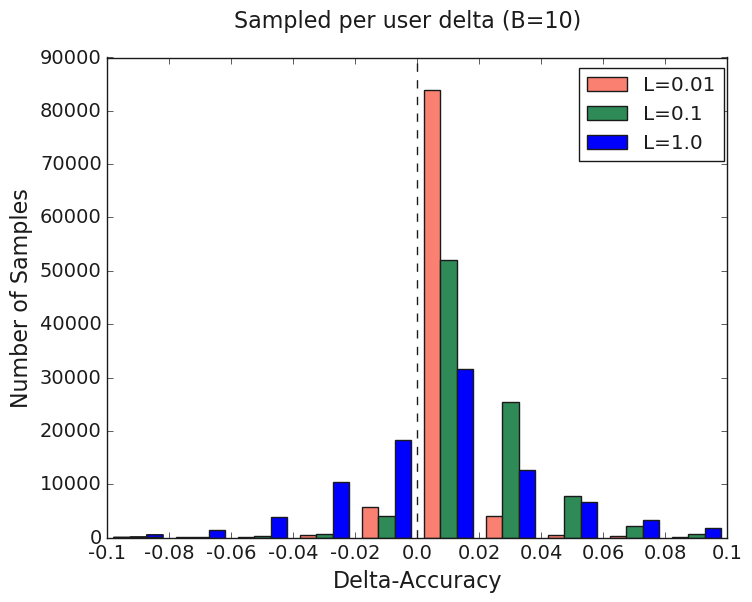}
    \caption{}
    \label{fig:1b}
  \end{subfigure}
  \begin{subfigure}{.32\linewidth}
    \includegraphics[width=\columnwidth]{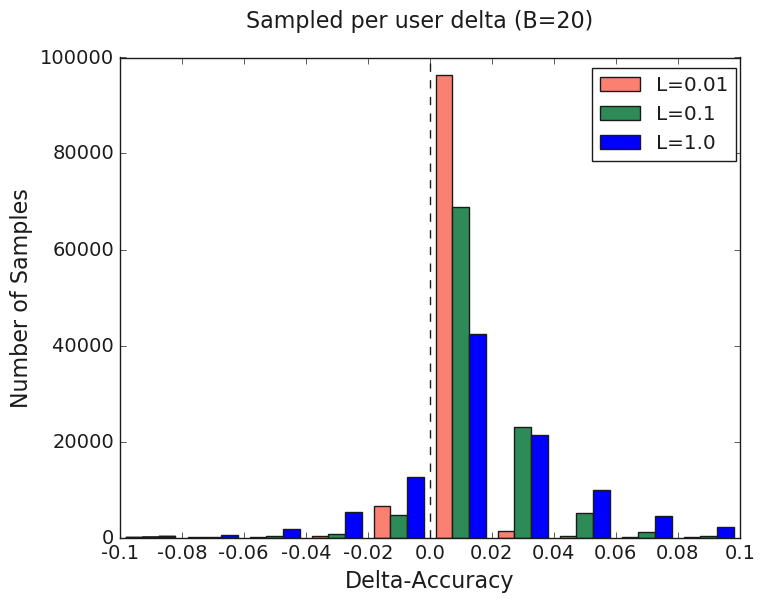}
    \caption{}
    \label{fig:1c}
  \end{subfigure}
  \caption{Accuracy delta histograms for different learning rates (L) and batch
    sizes (B): (a) At $B=5$ and $L=0.1$, 47\% of users achieve $\geq$ 0.02
    accuracy improvement; (b) At $B=10$ and $L=1.0$, 39\% of users
    achieve $\geq$ 0.02 accuracy improvement; (c) At $B=10$ and $L=0.1$,
    29\% of users achieve $\geq$ 0.02 accuracy improvement.}
  \vspace*{-4mm}
  \label{fig:fig1}
\end{figure*}

The best accuracy improvement is achieved for $B=5$, $L=0.1$. It starts
with a mean baseline prediction accuracy of 0.166 and reaches a mean
personalized accuracy of 0.19, resulting in a mean relative accuracy
increase of 14.5\%.

While the mean metrics show how much personalization improves the model
performance in general, the distribution reveals how personalization influences
the experience of individual users. Histograms of the sampled accuracy deltas
are shown in Figure~\ref{fig:fig1}.

As shown in Figure~\ref{fig:1a}, with a small batch size, a large
portion of users encounter model degradation with learning rate 1.0.
This is not entirely surprising, since high learning rates with
small batch size can cause the parameter update to jump over or even
divert from the minima. In Figure~\ref{fig:1b} and
Figure~\ref{fig:1c}, with larger batch sizes, histograms of learning
rate 1.0 tend to have heavier tails both on the left and on the right,
compared with histograms of learning rate 0.1. Though the average
accuracy improvement of learning rate 1.0 in batch size 20 is lower
than learning rate 0.1 (0.015 vs. 0.018), neither is clearly superior,
since more users (39\% vs. 29\%) achieve significant accuracy improvement
($\geq0.02$) with learning rate 1.0.

\begin{wrapfigure}{r}{0.6\textwidth}
  \centering
  \begin{subfigure}{0.29\textwidth}
    \includegraphics[width=\linewidth]{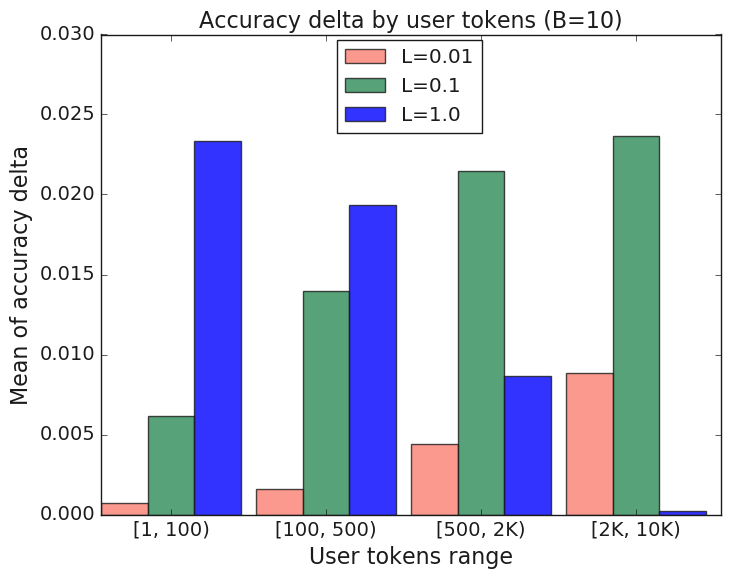}
    \caption{}
    \label{fig:2a}
  \end{subfigure}
  \begin{subfigure}{0.29\textwidth}
    \includegraphics[width=\linewidth]{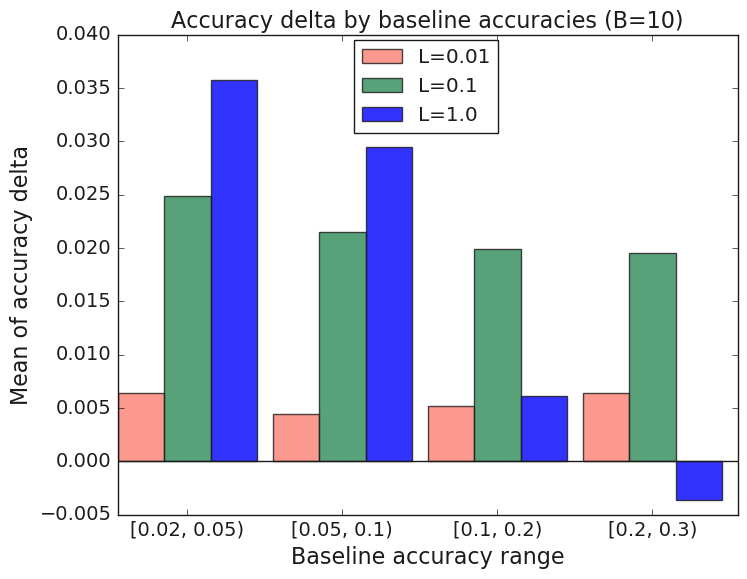}
    \caption{}
    \label{fig:2b}
  \end{subfigure}
  \caption{Analysis of accuracy deltas by learning rate (L) and batch
    size (B) sliced by (a) number of user tokens and (b) baseline accuracy.}
  \label{fig:fig2}
\end{wrapfigure}

All personalization evaluation experiments in our study use the data
stored in a user's on-device training cache. Variations in the
quantity and quality of training cache data across different devices
are expected. We conduct experiments to evaluate how model
personalization can be influenced by factors associated with the user
data. Factors considered are the number of training tokens and
baseline accuracy on the user data. The stopping criteria are set to
$T \geq 10000$ or $E \geq1$. Summaries of the greater improvement are
illustrated in Figure~\ref{fig:fig2}.

In Figure~\ref{fig:2a}, user token counts are placed into 4
buckets. As one might expect, we observe larger improvements for more
data. For a learning rate of 0.1, the accuracy improvements of the
last two buckets get closer, indicating the saturation of the
improvement. A learning rate of 1.0 retrieves the best performance
with very few tokens. The graph suggests that adjusting the learning
rate based on number of user tokens leads to better results.

In Figure~\ref{fig:2b}, baseline accuracies of users are placed into
4 buckets. With learning rates 0.1 and 1.0, accuracy improvement
for users with the worst baseline accuracy ($\leq 0.1$) is greater
than 0.25, while accuracy improvement for users with best baseline
accuracy ($\geq 0.2$) is smaller than 0.2. The results indicate
that users who deviate the most from the global model predictions are
those benefiting the most from it.

\section{Conclusion}

This work describes tools to perform Federated Personalization
Evaluation and analyze results in a privacy-preserving manner. Through
experiments on live traffic, we show that personalization benefits
users across a large population. We explore personalization strategies
and demonstrate how they can be tuned to achieve better
performance. To our knowledge, this represents the first evaluation of
personalization using privacy-preserving techniques on a large
population of live users.

\subsubsection*{Acknowledgments}

The authors would like to thank colleagues on the \assistant{} and
\googleai{} teams for many helpful discussions. We're especially
grateful to \eglanz{} and \mcmahan{} for help with our experiments,
and \harda{} for help editing the manuscript. We're also grateful
for the many contributions made by researcher Jeremy Kahn to an early
iteration of this work.

\small

\bibliography{neurips_2019}
\bibliographystyle{acl_natbib}

\end{document}